\documentclass[runningheads]{llncs}

\usepackage[mobile]{eccv}

\usepackage{eccvabbrv}
\usepackage{amsmath}
\usepackage{algorithm}
\usepackage{algpseudocode}
\usepackage{graphicx}
\usepackage{booktabs}
\usepackage[accsupp]{axessibility}  

\usepackage[pagebackref,breaklinks,colorlinks]{hyperref}

\usepackage{orcidlink}

\definecolor{best}{rgb}{1,0,0} 
\definecolor{secondbest}{rgb}{0,0,1} 
\definecolor{topthree}{gray}{0.8} 

\newif\ifdraft
\drafttrue

\ifdraft
\newcommand{\zwc}[1]{{\color{magenta}[\textbf{ZW:} \textit{#1}]}}

\newcommand{\jw}[1]{{\color{black}#1}}

\else
\newcommand{\zwc}[1]{}
\newcommand{\yzc}[1]{}

\newcommand{\jw}[1]{{\color{black}#1}}

\fi

\begin{document}

\title{DiffBody: Human Body Restoration by Imagining with Generative Diffusion Prior}

\titlerunning{DiffBody}

\author{
    Yiming Zhang\inst{1}\thanks{First author, \textsuperscript{†} Co-corresponding author.} \and
    Lionel Z. Wang\inst{2} \and
    Xinjie Li\inst{3} \and
    Yunchen Yuan\inst{2} \and\\
    Chengsong Zhang\inst{4} \and
    Xiao Sun\inst{5} \and
    Zhihang Zhong\inst{5}\textsuperscript{†} \and
    Jian Wang\inst{6}\textsuperscript{†}
}

\authorrunning{Zhang et al.}

\institute{Cornell University, Ithaca NY, USA \and
The Hong Kong Polytechnic University, Hong Kong SAR\and Pennsylvania State University, University Park PA, USA \and
University of Illinois Urbana-Champaign, Champaign IL, USA \and
Shanghai Artificial Intelligence Laboratory, Shanghai, China \and 
Snap Inc., New York NY, USA}

\maketitle

\begin{abstract}

Human body restoration plays a vital role in various applications related to the human body.
Despite recent advances in general image restoration using generative models, their performance in human body restoration remains mediocre, often resulting in foreground and background blending, over-smoothing surface textures, missing accessories, and distorted limbs. 
Addressing these challenges, we propose a novel approach by constructing a human body-aware diffusion model that leverages domain-specific knowledge to enhance performance. 
Specifically, we employ a pretrained body attention module to guide the diffusion model's focus on the foreground, addressing issues caused by blending between the subject and background.
We also demonstrate the value of revisiting the language modality of the diffusion model in restoration tasks by seamlessly incorporating text prompt to improve the quality of surface texture and additional clothing and accessories details.
Additionally, we introduce a diffusion sampler tailored for fine-grained human body parts, utilizing local semantic information to rectify limb distortions.
Lastly, we collect a comprehensive dataset for benchmarking and advancing the field of human body restoration.
Extensive experimental validation showcases the superiority of our approach, both quantitatively and qualitatively, over existing methods.

\end{abstract}

\section{Introduction}

Blind image restoration (BIR) aims to enhance the quality of degraded images through processes like denoising \cite{tian2020deep}, sharpening \cite{wang2020experiment}, deblurring \cite{zhang2022deep}, super-resolution \cite{liu2022blind}, \etc, a domain that has seen significant progress with advancements in the data-driven learning paradigm.
Although generalized BIR has made substantial strides, users often exhibit a greater interest in the specific effects of BIR on particular subjects, with the human body being one of the key focuses.
The restoration of the human body can have a profound impact on various human-centric applications, such as improving portrait quality in social media apps and aiding in related downstream tasks like ReID \cite{ye2021deep}, 3D reconstruction \cite{wang2021deep}, \etc.

\begin{figure}[t]
    \centering
    \includegraphics[width=\linewidth]{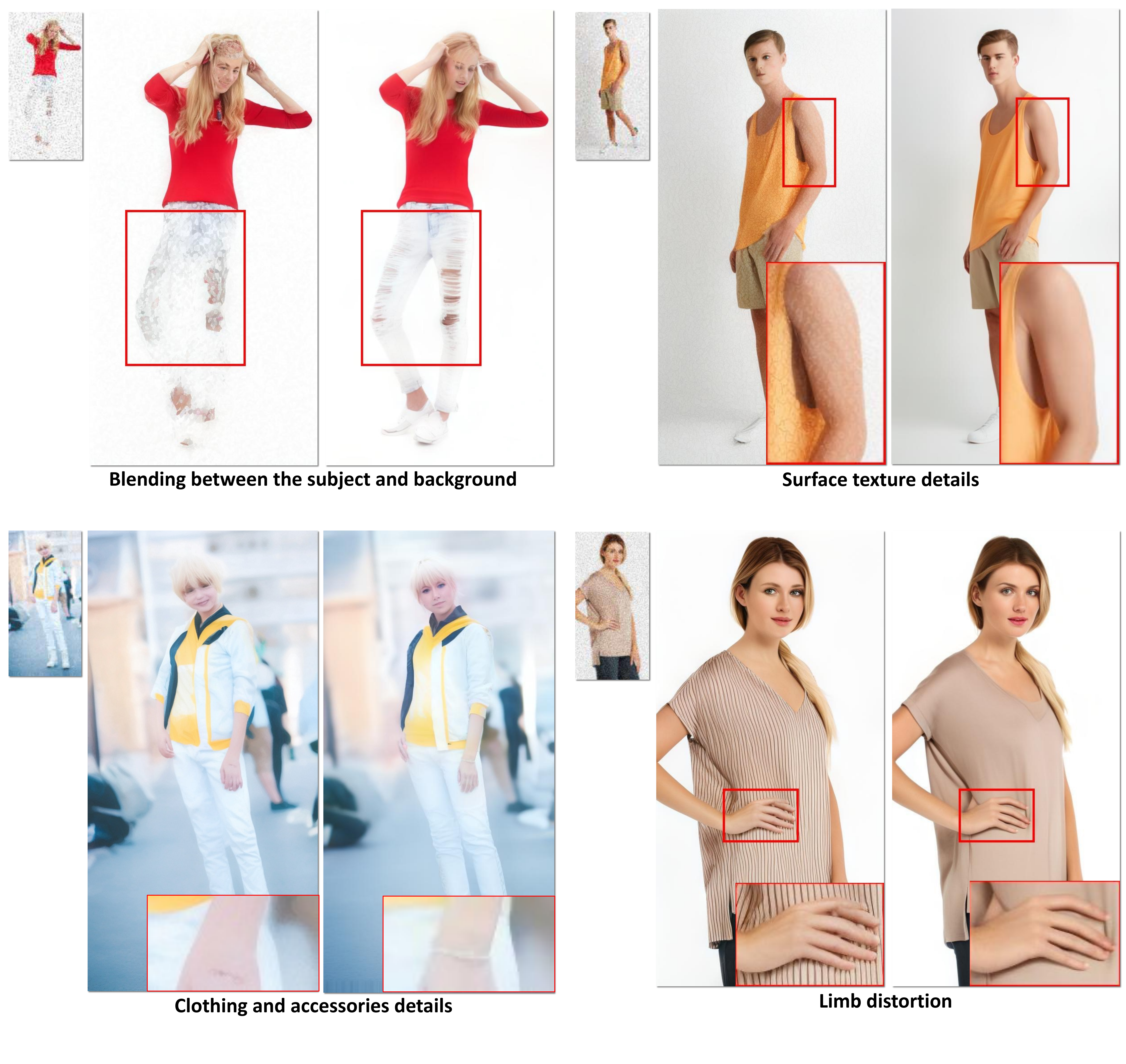}
    \vspace{-1cm}
    \caption{Comparison between our model and baseline (Left: Baseline, Right: Ours, Top left corner: LQ input). Comparing to baseline, our model has better performance on problems labeled below each image.}
    \label{fig:fig1}
\vspace{-0.5cm}
\end{figure}

Regarding the methodology of BIR, while the end-to-end reconstruction paradigm \cite{liang2021swinir, wang2021real} has made great progress, it struggles to handle complicated combinatorial and severe degradations. 
The generative paradigm offers a solution to this issue by harnessing the power of generative models, such as Generative adversarial networks (GANs) \cite{DBLP:journals/corr/abs-1812-04948} and Diffusion models \cite{DBLP:journals/corr/abs-2112-10752}.
The priors of generative models possess a powerful ``imagination'' learned from large amounts of data, which can be used to fill in reasonable details of degraded images.
Thus, current diffusion-based image restoration models \cite{luo2023refusion, lin2023diffbir} have notably enhanced the perceptual quality and adaptability of restoration results, thereby expanding the applicability of image restoration in practical contexts.

Despite these advancements, the specific area of human body image restoration remains underdeveloped. 
It is worth noting that even current diffusion-based general restoration models \cite{luo2023refusion, lin2023diffbir} may produce artifacts for low-quality human images, including foreground and background blending, over-smoothing surface textures, missing accessories, and distorted limbs, as illustrated in \cref{fig:fig1}.
These issues restrict the practical deployment of current models in real-world applications, highlighting the need for specialized solutions in the promising domain of human body image restoration.
However, the theoretical upper limit of performance for human body restoration is higher than that for general restoration, since existing knowledge of the human body can be utilized as additional information.
Therefore, how to specialize the generative prior-based general restoration model, which performs mediocrely on low-quality human body images, into the human body-aware restoration model, which performs excellently, is the key problem to be studied in this work.

The key idea of this work is to guide and mine the pretrained diffusion model to generate clear and realistic imagery of the human body.
We present DiffBody, a novel and specialized diffusion model designed specifically for human body image restoration.
First, we meticulously collect a high-quality human body dataset for benchmarking the human body restoration task.
Regarding the methodology, we combine pretrained restoration models from both the reconstruction paradigm, SwinIR \cite{liang2021swinir}, and the generative paradigm, ControlNet \cite{zhang2023adding}, to create a robust backbone.
To address the issue of foreground and background blending, we employ structural priors, including the pose and global attention map of the human body, to guide the diffusion model's focus towards the human body.
Additionally, we recognize the untapped potential of the diffusion model's understanding of language modality for human body restoration. 
By incorporating human-centric text guidance with detailed descriptions of human appearance, we enhance surface texture quality and add details of clothing and accessories.
Furthermore, we introduce a diffusion sampler tailored for fine-grained human body parts, utilizing local semantic information to rectify limb distortions.
These proposed techniques allow for the restoration of degraded images with an emphasis on human-centric features, ensuring the diffusion model has a clearer and more reasonable imaginary of the human body.
As illustrated in \cref{fig:fig1}, our method achieves significantly higher fidelity and quality compared to previous approaches, which suffer from various artifact issues.

Our main contributions are as follows: 
1) We introduce a new dataset of human images and achieve state-of-the-art human body image restoration performance for benchmarking the human body restoration task;
2) We utilize the power of pose and global attention map guidance, text guidance, and a human-centric diffusion sampler to address common artifacts encountered by general diffusion-based restoration models, such as foreground and background blending, over-smoothing surface textures, and distorted limbs;
3) We systematically verify the effectiveness and rationality of our proposed techniques through extensive experiments, including quantitative comparison, qualitative comparison, user studies, and ablation studies.

\section{Related Work}

\subsection{Blind Image Restoration}

Blind image restoration (BIR) aims to enhance image restoration efforts by leveraging a pre-trained prior network within an unsupervised framework. Predominantly, existing literature has concentrated on discerning a latent code situated in the latent space of a pre-trained Generative Adversarial Network (GAN), as evidenced by works such as \cite{bora2017compressed,daras2021intermediate,menon2020pulse,pan2021exploiting}. Recent advancements in this domain have transitioned towards the utilization of Denoising Diffusion Probabilistic Models (DDPMs), as delineated in seminal studies by \cite{ho2020denoising,song2019generative,song2020score,rombach2022high,ramesh2022hierarchical,saharia2022photorealistic}, marking a notable shift from conventional approaches. Other novel approaches such as DDRM \cite{kawar2022denoising} utilizes Singular Value Decomposition (SVD) to address linear image restoration challenges, presenting an innovative and simplified approach and DDNM \cite{wang2022zero} delves into vector range-null space decomposition to develop a novel sampling strategy, enhancing image restoration efficiency. In the realm of domain-specific image restoration models, a predominant emphasis has been placed on blind face restoration, as evidenced by works such as \cite{liu2022blind, wang2022restoreformer, gu2022vqfr}. In contrast, the equally critical domain of human body restoration has not seen comparable development, a gap that our DiffBody model seeks to address.

\vspace{-3mm}
\subsection{Controllable Human Image Generation}

Traditional methods for generating controllable human images mainly fall into two categories: those based on Generative Adversarial Networks (GANs) \cite{zhu2017your,siarohin2019appearance} and those using Variational Autoencoders (VAEs) \cite{ren2020deep,yang2021towards}, both leveraging reference images and specific conditions for input. Recent studies have ventured into enabling the generation process through textual instructions, though these tend to limit user input to basic pose or style adjustments \cite{roy2022tips,jiang2022text2human}. Cutting-edge developments for creating human images with detailed control over vocabulary and pose include ControlNet\cite{zhang2023adding}, T2I-Adapter\cite{mou2023t2i}, HumanSD\cite{ju2023humansd}, and HyperHuman\cite{liu2023hyperhuman}. These works have shown that diffusion models are capable to generate human image contains rich detail and natural texture, which we believe can be utilized for our human body image restoration model. 

\vspace{-3mm}
\subsection{Datasets for Human Image Generation}

The compilation of extensive datasets is crucial for the advancement of image generation technologies, particularly within the realm of human image generation. Datasets such as iDesigner \cite{dufour2022scam}, Market1501 \cite{zheng2015scalable}, DeepFashion \cite{liu2016deepfashion}, and MSCOCO \cite{lin2014microsoft} have historically been centered on generating images of real-scene humans, providing noisy paired source-target imagery. The LIP dataset \cite{gong2017look} comprises approximately 50,000 human-centric images, filled with comprehensive annotations that include nineteen semantic human part labels in addition to one for the background. Human-Art \cite{ju2023human} offers a corpus of 50,000 human-centric images across a blend of five natural and fifteen artificial scenes, each annotated with precise pose data and text. VITON \cite{han2018viton} is focused on human-clothing pairing, while SHHQ \cite{fu2022stylegan} primarily showcases full-bodied humans against clean backgrounds, with an emphasis on fashion imagery that exhibits minimal pose variation. Despite the public accessibility and richness in text-image pairs of the LAION-5B dataset \cite{schuhmann2022laion}, it includes a significant quantity of content irrelevant to human images. ControlNet \cite{zhang2023adding} utilizes the human pose estimator OpenPose \cite{cao2017realtime} to collect a dataset of 200,000 pose-image-text triplets from internet-sourced images, the majority of which depict real-world scenes. HumanSD \cite{ju2023humansd} introduced two datasets; GHI \cite{ju2023humansd} encompasses one million multi-scenario images synthesized from SD using meticulously formulated prompts, whereas LAION-Human \cite{ju2023humansd} distills one million human-centric images from the broader LAION-5B dataset through a stringent filtering process. Surpassing these, HumanVerse \cite{liu2023hyperhuman} presents a collection of 340 million images, each annotated with an array of details including human pose, depth, and surface normals. In this research, we introduce the Web-human dataset, aiming to augment the repository of available resources by amassing an additional 500,000 high-resolution human images sourced from web photography albums and portrait collections. We train our model, DiffBody, on a hybrid dataset comprised of SHHQ \cite{fu2022stylegan}, DeepFashion \cite{liu2016deepfashion}, and our novel Web-human dataset, resulting in a consolidated dataset of 140,000 images, each accompanied by exhaustive descriptive annotations. This expanded dataset landscape significantly enhances the capabilities of HIG models, enabling the generation of more nuanced human images.

\vspace{-3mm}
\section{Methodology}
\vspace{-3mm}

\subsection{Preliminary}

\textbf{ControlNet}: ControlNet\cite{zhang2023adding} is an advanced neural network framework designed to enhance text-to-image diffusion models by incorporating specific image conditions. Given an input image $z_0$, image diffusion algorithms progressively add noise to the image, producing a noisy image $z_t$, where $t$ signifies the number of noise addition iterations. ControlNet introduces a set of conditions, including the time step $t$, text prompts $c_t$, and a task-specific condition $c_f$. These algorithms learn a network $\epsilon_\theta$ to predict the noise added to the noisy image $z_t$. The learning objective $L$, integral to the entire diffusion model's optimization, is expressed as:

\begin{equation}
L(\theta) = \mathbb{E}_{z_0, \epsilon, t, c_t, c_f} \left[ \| \epsilon - \epsilon_\theta(z_t, t, c_t, c_f) \|_2^2 \right]
\end{equation}

  This equation represents the expected difference between the actual noise $\epsilon$ and the noise predicted by the network $\epsilon_\theta$, given the conditions at each time step. The objective $L$ is directly utilized in fine-tuning diffusion models with ControlNet, aiming to minimize this difference and thus enhance the fidelity and relevance of the generated images to the given conditions.

\begin{figure}[htbp]
    \centering
    \includegraphics[width=\linewidth]{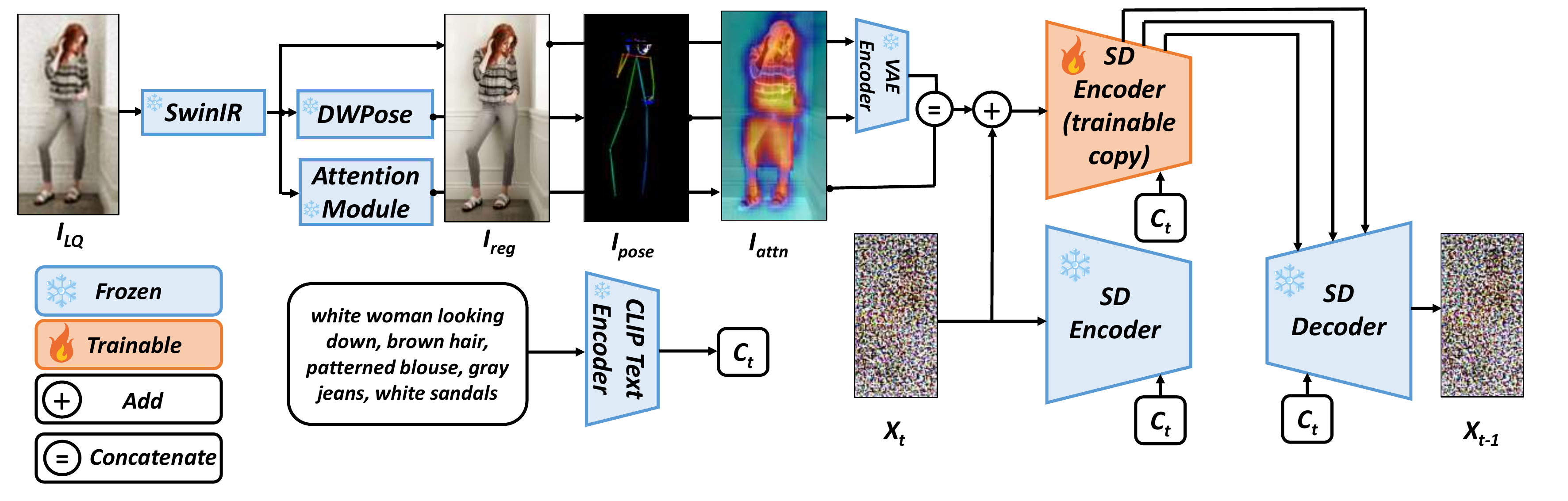}
    \caption{The structure of DiffBody. First, we train the SwinIR model using our proposed dataset and process the low-quality image $I_{LQ}$ to obtain preliminary restored image $I_{reg}$ with the trained model. In addition, pose map $I_{pose}$ and attention map $I_{attn}$ are extracted from $I_{reg}$ using existing methods. Afterwards, $I_{reg}$ and $I_{pose}$ are passed into the pre-trained VAE Encoder, then concatenated together with $I_{attn}$ and \jw{fed} to the trainable copy of SD Encoder. Additionally, we also utilize the textual information (Sec 3.3) and a novel human-centric sampling (Sec 3.4) to enhance the restoration capability. Please see corresponding sections for details.} 
    \label{fig:enter-label}
\vspace{-0.2cm}
\end{figure}

\subsection{Enhancing Human Image Restoration through Structural Guidance}

In developing a robust pipeline for human image restoration, we initially target the reduction of degradations observable in low-quality (LQ) images. This foundational step ensures that subsequent processing stages can more effectively discern features within these images without interference from existing impairments. To achieve this, we incorporate the SwinIR \cite{liang2021swinir} model architecture, which has been pre-trained on a dataset pertinent to our domain of interest and further refined through fine-tuning on our specialized dataset for human bodies. The primary goal of the restoration module's optimization revolves around minimizing the $L_2$ pixel loss, described mathematically as:

\begin{equation}
I_{\text{reg}} = \text{SwinIR}(I_{\text{LQ}}), \quad
L_{\text{reg}} = \|I_{\text{reg}} - I_{\text{HQ}}\|^2_2
\end{equation}
\jw{where} $I_{\text{HQ}}$ and $I_{\text{LQ}}$ represent the high-quality and low-quality images, respectively, \jw{and} $I_{\text{reg}}$, the output of regression learning, \jw{is} set to undergo further restoration processes.

A notable challenge encountered with $I_{\text{reg}}$ includes its propensity towards oversmoothing and detail loss—typical artifacts of conservative image restoration methods. However, the efficacy of SwinIR in noise reduction enables subsequent pose detection and attention detection models to operate effectively on $I_{\text{reg}}$. Consequently, we employ both a body pose detection model\cite{yang2023effective} and a body part attention model \cite{somers2023body} to generate, respectively, the pose and attention maps for the human body:
\begin{equation}
I_{\text{pose}} = \text{DWPose}(I_{\text{reg}}), \quad
I_{\text{attn}} = \text{Attn}(I_{\text{reg}})
\end{equation}
In this framework, $I_{\text{pose}}$ refers to the pose image derived from $I_{\text{reg}}$, while $I_{\text{attn}}$ captures the attention heatmap of the human body as discerned from $I_{\text{reg}}$. This innovative approach underscores our commitment to enhancing human image restoration through the integration of structural guidance, effectively addressing common restoration challenges while setting the stage for more nuanced and detail-rich reconstructions.

\subsection{Leveraging Textual Information for Image Restoration}
\begin{figure}
\vspace{-0.25cm}
\centering
\includegraphics[width=\linewidth]{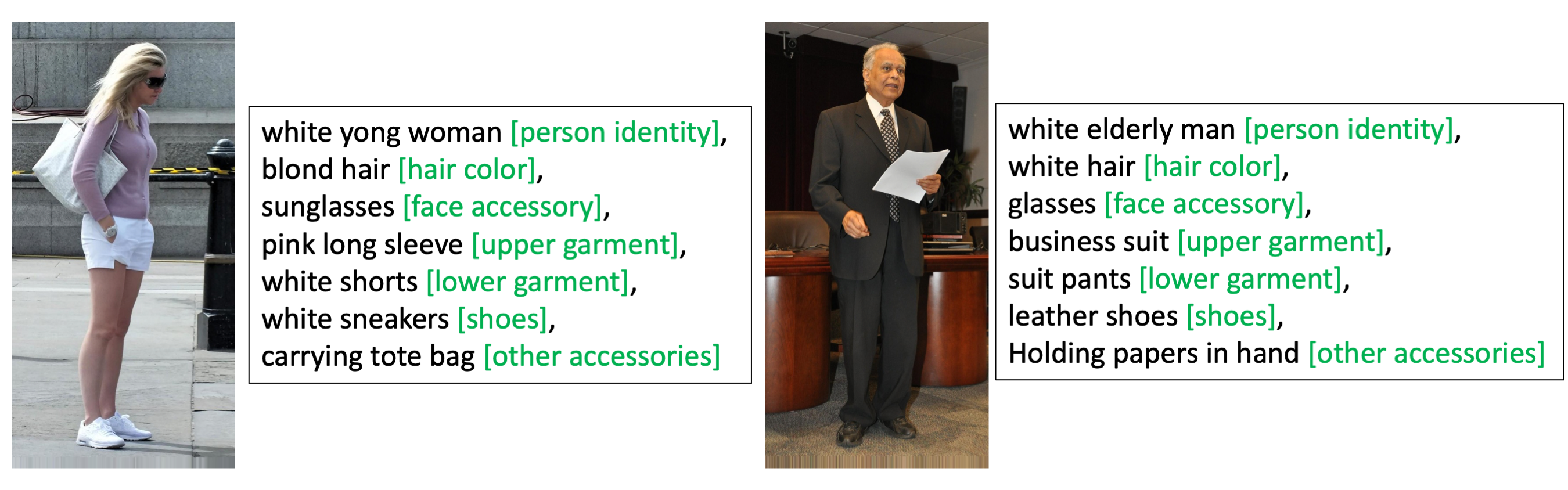}
\caption{During training, texts in black are fed to the model. Texts in green reflect the generative logic of GPT-4V in captioning images.}
\label{fig:prompt}
\vspace{-0.25cm}
\end{figure}

Conventional image restoration models have largely overlooked the utilization of textual information, which represents a significant and untapped prior knowledge source. This oversight overlooks the potential of text to significantly enhance the generation of high-quality images. In our approach, we harness uniformly formatted textual descriptions during the training phase of the latent diffusion model, specifically designed for human-centric subjects. By employing the GPT-4V model \cite{openai2024gpt4}, we generate detailed descriptions of high-quality human images, adhering to a meticulously defined sequence from top to bottom. During the inference phase, these structured textual prompts markedly improve the model’s precision in reconstructing images. \cref{fig:prompt} provides illustrative examples of the uniformly formatted textual prompts utilized.

Upon establishing the foundational requirements for training, we characterize the latents of $I_{\text{reg}}$ and $I_{\text{pose}}$, as encoded by the Variational Autoencoder (VAE), with $\mathcal{E}(I_{\text{reg}})$ and $\mathcal{E}(I_{\text{pose}})$ respectively (For detailed information, refer to \cref{fig:enter-label}). The defined learning objective, which guides our model training, is as follows:
\begin{equation}
L_{\text{Diff}}(\theta) = \mathbb{E}_{z_t, t, c_t, \epsilon(I_{\text{reg}})} \left[ \left\| (1+I_{\text{attn}}) \cdot(\epsilon - \epsilon_{\theta}(z_t, t, c_t, (\mathcal{E}(I_{\text{reg}}), \mathcal{E}(I_{\text{pose}}), I_{\text{attn}}))) \right\|_2^2 \right].
\end{equation}
The essence of this strategy is to broaden the model's learning paradigm to incorporate not only the degraded images but also textual descriptions, poses, and attention maps. This comprehensive approach is devised to augment the restoration process by integrating varied data sources, a critical factor for the meticulous recovery of details lost to image degradation. By synergizing these diverse yet complementary inputs, our model is adept at restoring images with enhanced fidelity, especially in instances where critical information has been obscured.

\vspace{-2mm}
\subsection{Human-centric Guidance for Diffusion Sampling}
\vspace{-2mm}

Despite the commendable restoration outcomes achieved by our aforementioned strategy, challenges persist during the diffusion process in the Latent Diffusion Model (LDM), particularly artifacts synonymous with human generation, such as limb distortions and the omission of finer details like fingers. Prior research efforts dedicated to refining the quality of human limbs in image generation have necessitated the use of accurate human control maps, such as depth or mesh maps, which are notably challenging to derive from $I_{\text{reg}}$. In response to these challenges, innovative approaches\cite{avrahami2022blended, fei2023generative} have guided the intermediate variable $x_0$ to influence the generation process of diffusion models. Drawing inspiration from these endeavors, our objective is to derive a pristine latent representation $z_0$ utilizing the following equation:
\begin{equation}
\tilde{z}_0 = \frac{z_t}{\sqrt{\bar{\alpha}_t}} - \frac{\sqrt{1 - \bar{\alpha}_t} \epsilon_{\theta}(z_t, t, c_t, (\mathcal{E}(I_{\text{reg}}), \mathcal{E}(I_{\text{pose}}), W_a))}{\sqrt{\bar{\alpha}_t}}
\end{equation}
  The clean latent representation $z_0$ is subsequently decoded using the VAE decoder to produce the restored image, denoted as $\mathcal{D}(\tilde{z}_0)$. Further, we leverage the architecture of a body part-based Re-Identification (ReID) model \cite{somers2023body} to extract features from both $I_{\text{reg}}$ and $\mathcal{D}(\tilde{z}_0)$, facilitating the computation of a body-part based loss $L_{\text{part}}$:
\begin{equation}
L_{\text{part}}(x, I_{\text{reg}}) = \mathcal{L}(\mathcal{D}(\tilde{z}_0), I_{\text{reg}}) = \sum_i L_{CE} \left(\mathcal{H_\text{i}}(\mathcal{D}(\tilde{z}_0)) - \mathcal{H_\text{i}}(I_{\text{reg}}) \right)
\end{equation}
  Here, $\mathcal{H}$ signifies the body part feature extractor, targeting key human components such as the head, torso, hands, legs, and feet. This loss mechanism iteratively ensures semantic alignment between $I_{\text{reg}}$ and $\mathcal{D}(\tilde{z}_0)$ throughout the inference sampling process, effectively retaining critical body part information that the LDM model might otherwise misinterpret. The comprehensive algorithm for our latent image guidance is encapsulated in Algorithm \ref{alg:alg}.

\begin{algorithm}[htbp]

\caption{Employing a diffusion model $\epsilon_{\theta}$ alongside the VAE's encoder $\mathcal{E}$ and decoder $\mathcal{D}$ for enhanced image restoration.}
\begin{algorithmic}[1]
\Require Guidance image $I_{\text{reg}}$, text description $c_t$, diffusion steps $T$, gradient scale $s$, body part feature extractor backbone $\mathcal{H}$
\Ensure Output image $D(\tilde{z}_0)$
\State Sample $x_T$ from $\mathcal{N}(0, I)$
\For{$t$ from $T$ to $1$}
    \State $\tilde{z}_0 \leftarrow \frac{z_t}{\sqrt{\bar{\alpha}_t}} - \frac{\sqrt{1 - \bar{\alpha}_t} \epsilon_{\theta}(z_t, t, c_t, (\mathcal{E}(I_{\text{reg}}), \mathcal{E}(I_{\text{pose}}), I_{attn}))}{\sqrt{\bar{\alpha}_t}}$
    \State $\mathcal{L} \leftarrow \mathcal{L_{\text{part}}}(\mathcal{D}(\tilde{z}_0), I_{\text{reg}})$
    \State Sample $\tilde{z}_{t-1}$ by $\mathcal{N}(\mu_{\theta}(\tilde{z}_t) - s \nabla_{\tilde{z}_t}\mathcal{L}, \sigma_t^2)$
\EndFor
\State \Return $D(\tilde{z}_0)$
\end{algorithmic}
\label{alg:alg}
\end{algorithm}
\vspace{-3mm}
\section{Experiment}
\vspace{-3mm}
\subsection{Datasets}

Training diffusion models for image restoration necessitates comprehensive datasets, covering a broad spectrum of image scenarios to ensure a diverse training environment. High data quality is paramount, particularly for image restoration applications where precision and fidelity are crucial. Given the prevalent challenges in current human image datasets and limitations of web crawlers—ranging from missing to poorly represented human figures—we have implemented an exhaustive dataset development protocol. This process is meticulously structured into four phases to address these concerns effectively: (1) initial screening based on image format, (2) detection of human figures within the images, (3) adjustment of the output format to match our model's requirements, and (4) a thorough manual inspection to guarantee the utmost quality.\\
\textbf{SHHQ}: The SHHQ dataset \cite{fu2022stylegan} is a collection of high-quality, full-bodied human images rendered at a resolution of 1024 × 512, with subjects centrally positioned to facilitate the generation of accurate pose, attention maps, and textual descriptions. However, its primary drawback is the limited diversity in poses, especially more complex ones, with only 40,000 images available—insufficient for our comprehensive training needs. Through our rigorous cleaning process, we refine this dataset for enhanced utility.\\
\textbf{DeepFashion}: DeepFashion \cite{liuLQWTcvpr16DeepFashion}, a vast repository of fashion imagery, encompasses over 800,000 images ranging from professional shoots to everyday consumer photographs. This diversity presents a valuable resource for sourcing high-quality human images. Despite its extensive collection, a significant portion of the images do not meet our specified quality criteria. Our tailored cleaning methodology is employed to select and retain 50,000 images that adhere to our high-quality standards for model training.\\
\textbf{Web-sourced Images}: Beyond SHHQ and DeepFashion, we have curated 500,000 high-resolution images from web photography and portrait collections, offering a rich variety of scenes and poses. Following our data cleaning protocol, this pool has been refined to 50,000 premium human images.\\
Our comprehensive training dataset for the DiffBody model amalgamates the refined selections from SHHQ, DeepFashion, and web-sourced collections, culminating in a total of 140,000 meticulously described images. For evaluation purposes, a subset comprising 2,000 images was randomly selected to form our test dataset. This extensive and diverse dataset underpins the robust training of our diffusion model, ensuring its adeptness at image restoration across a wide array of human images.
\vspace{-4mm}
\subsection{Experimental Details}
\vspace{-2mm}
\textbf{Detailed Implementation.}
We employ Stable Diffusion 2.1-base \cite{rombach2022high} as our primary generative backbone, fine-tuning the DiffBody model across 10,000 iterations with a batch size of 128. Optimization is achieved through the Adam optimizer at a learning rate of $10^{-4}$, executed at 1024 × 512 resolution on four NVIDIA A100 GPUs. During inference, we leverage DDPM sampling over 50 steps. The DiffBody model adeptly handles images exceeding 1024 × 512 dimensions, recommending upsampling for smaller inputs to maintain peak efficiency. This method significantly bolsters our model's proficiency in generating and restoring images at high resolutions.
 \\
\textbf{Evaluation Metrics.}
In evaluating against ground truth, we utilize conventional metrics: PSNR, SSIM, and LPIPS \cite{zhang2018unreasonable}. To more accurately assess image authenticity for the BIR task, we incorporate no-reference image quality assessment (IQA) metrics: MANIQA \cite{yang2022maniqa} and CLIPIQA \cite{wang2023exploring}, enhancing our evaluation framework.

\vspace{-4mm}
\subsection{Comparisons with State-of-the-Art Methods}
\vspace{-2mm}

\begin{figure}[htbp]
    \centering
    \captionsetup{skip=5pt}
    \includegraphics[width=0.9\textwidth]{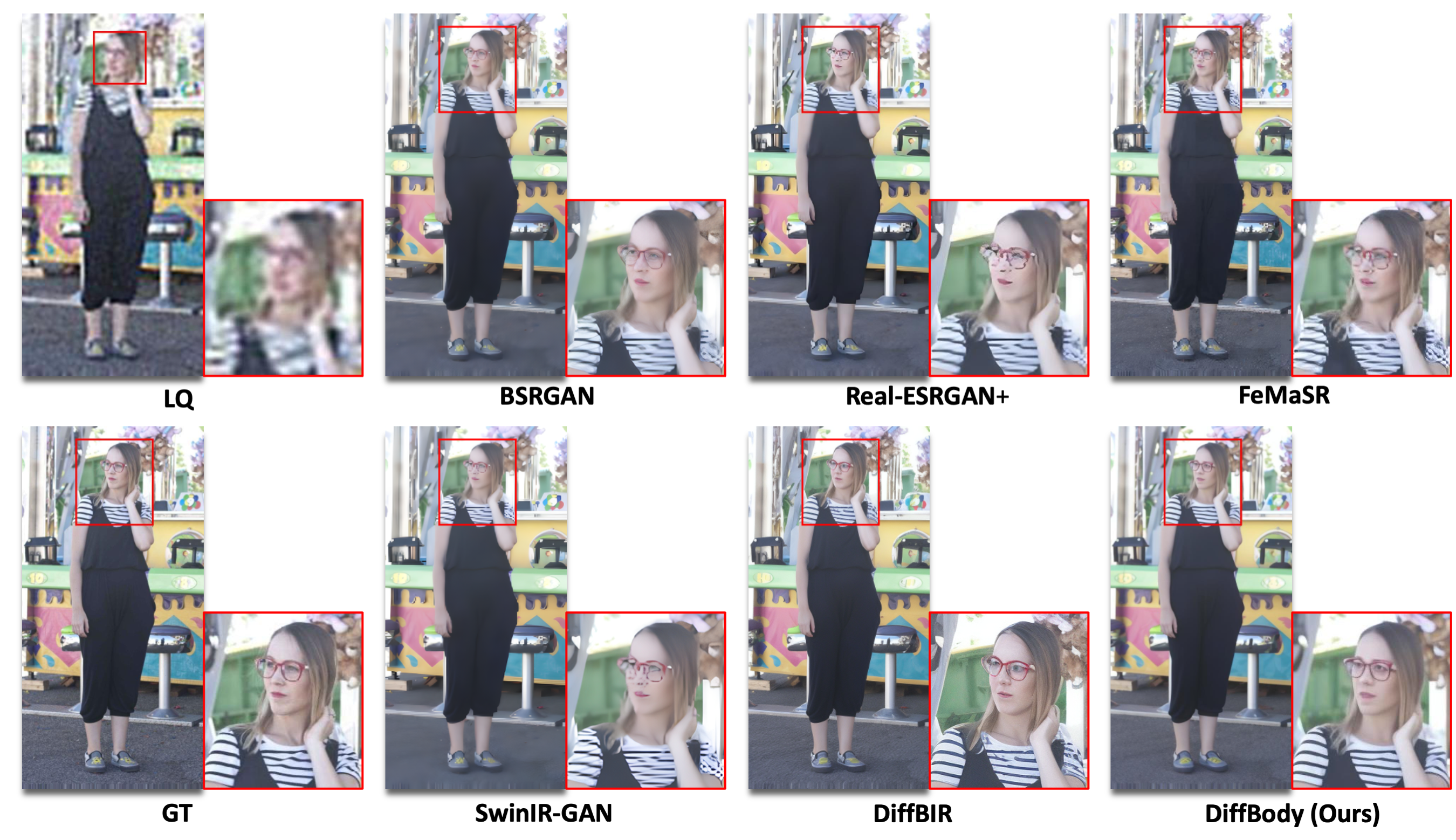}
    \caption{Visual comparison of DiffBody and other general SOTA methods. Compared to other methods, our model is more effective to generate reasonable human face and high-quality clothing.}
    \label{fig:sota1}
\end{figure}

\begin{figure}[htbp]
    \centering
    \includegraphics[width=0.9\textwidth]{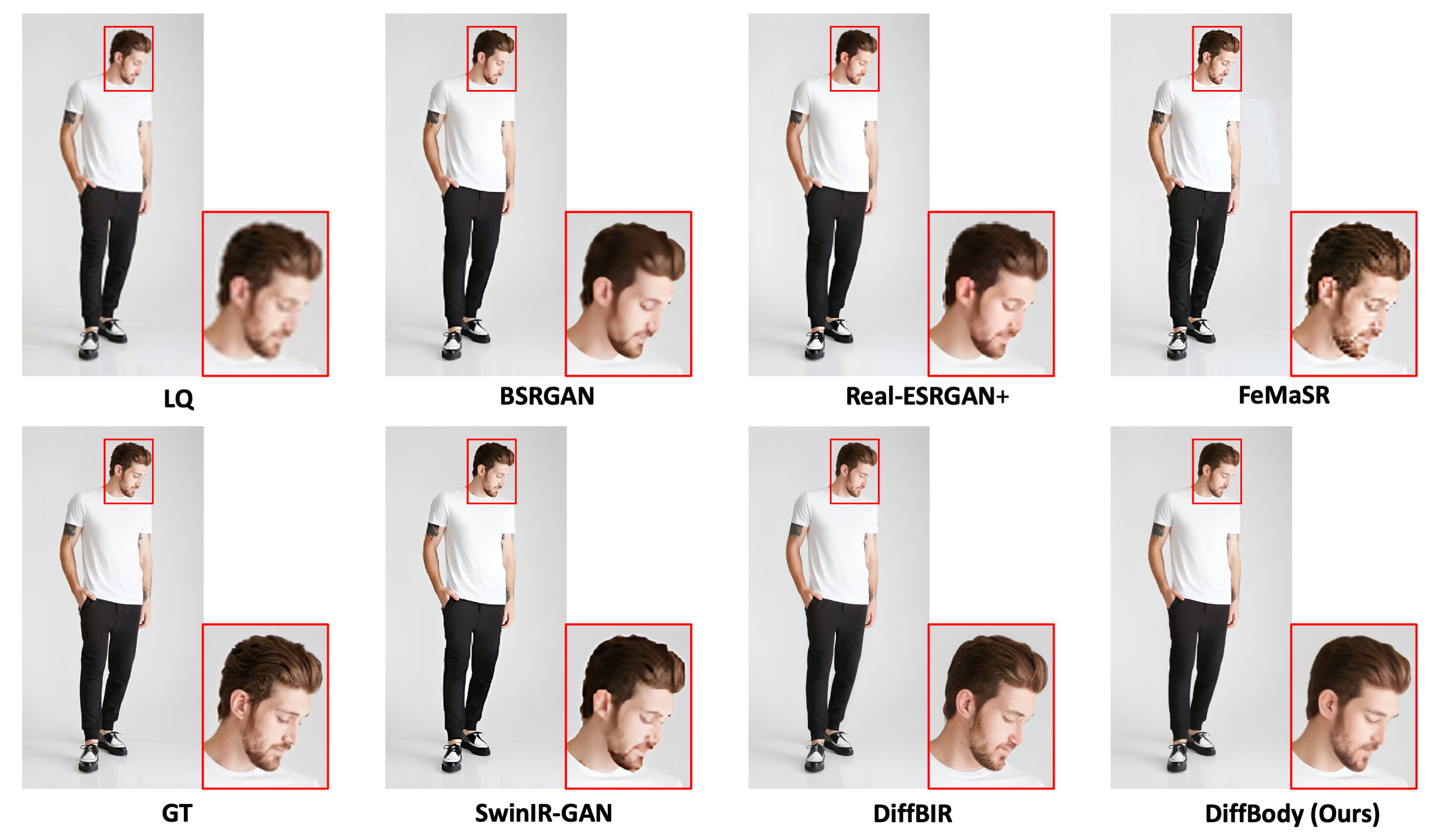}
    \caption{Visual comparison of DiffBody and other general SOTA methods. Compared to other methods, our model is more effective on generating natural human skin texture. (Zoom in to find the oil-painting-like artifacts in the result of DiffBIR.)}
    \label{fig:sota2}
\end{figure}

\begin{figure}[t!]
    \centering
    \includegraphics[width=0.8\textwidth]{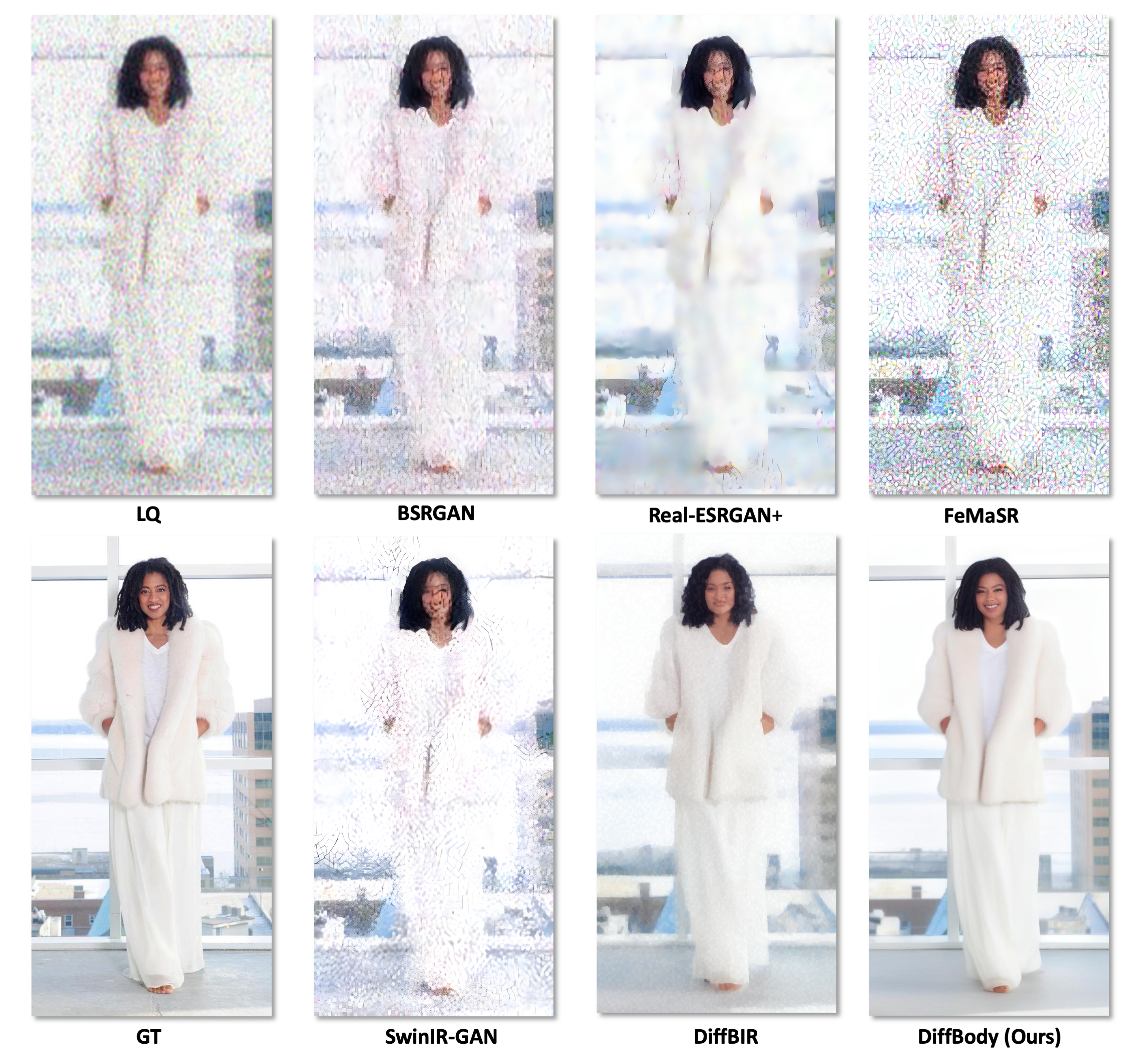}
    \caption{Visual comparison of DiffBody and other methods under extremely degraded cases (coupled degradation including adding noise, blur, and JPEG compression).}
    \label{fig:sota3}
\end{figure}

\noindent \textbf{Qualitative comparisons.} 
In the domain of human body restoration, we compare DiffBody with leading general image restoration methods: BSRGAN\cite{zhang2021designing}, Real-ESRGAN+\cite{wang2021real}, SwinIR-GAN\cite{liang2021swinir}, FeMaSR\cite{chen2022real}, and DiffBIR\cite{lin2023diffbir}. As evidenced by \cref{fig:sota1}, our model exhibits superior facial detail restoration capabilities relative to competing approaches. Furthermore, \cref{fig:sota2} demonstrates our model's unparalleled proficiency in reconstructing human body textures. \cref{fig:sota3} highlights our model's unique ability to reconstruct coherent human body imagery under severely degraded conditions, surpassing other models in this aspect.

\begin{table}[h!]
\centering
\caption{Quantitative comparisons with other methods. Best results are labeled in \textcolor{blue}{blue} and second best results are labeled in \textcolor{red}{red}.}
\label{table}
\setlength{\tabcolsep}{6pt}
\begin{tabular}{c c c c c c}
\toprule
Method & PSNR↑ & SSIM↑ & LPIPS↓ &  MANIQA↑ & CLIPIQA↑ \\ \midrule
BSRGAN &  22.2010 &0.5779  & 2.7035 & 0.3302 & 0.7423 \\ 
ESRGAN &  22.3916 & 0.6466  & 2.6448 &  0.3979 & 0.7422 \\ 
SwinIR-GAN & 21.2867 & 0.5750 & 2.6320 &  0.3369 & 0.7398  \\ 
FeMaSR & 21.2203 & 0.4849 & 2.8061  &   0.4189 & 0.7281 \\ 
DiffBIR &  \textcolor{blue}{24.1738}&  \textcolor{red}{0.7430}& \textcolor{red}{2.2433} &  \textcolor{red}{0.5300} & \textcolor{red}{0.7577} \\ 
Ours &  \textcolor{red}{24.0986} &  \textcolor{blue}{0.8446} &  \textcolor{blue}{1.8627} &   \textcolor{blue}{0.7222} &  \textcolor{blue}{0.7630} \\ \bottomrule
\end{tabular}
\end{table}

\vspace{-7mm}
\noindent \textbf{Quantitative comparisons}. We present a quantitative analysis based on 100 randomly selected images from our test dataset. According to \cref{table}, our model secures the top position in conventional benchmarks such as SSIM and LPIPS, while ranking as the second best in terms of PSNR. Regarding no-reference image quality metrics, our model outperforms all competitors on both MANIQA and CLIPIQA assessments.

\begin{figure}[t!]
\centering
\begin{subfigure}{\linewidth}
  \centering
  \includegraphics[width=0.8\linewidth]{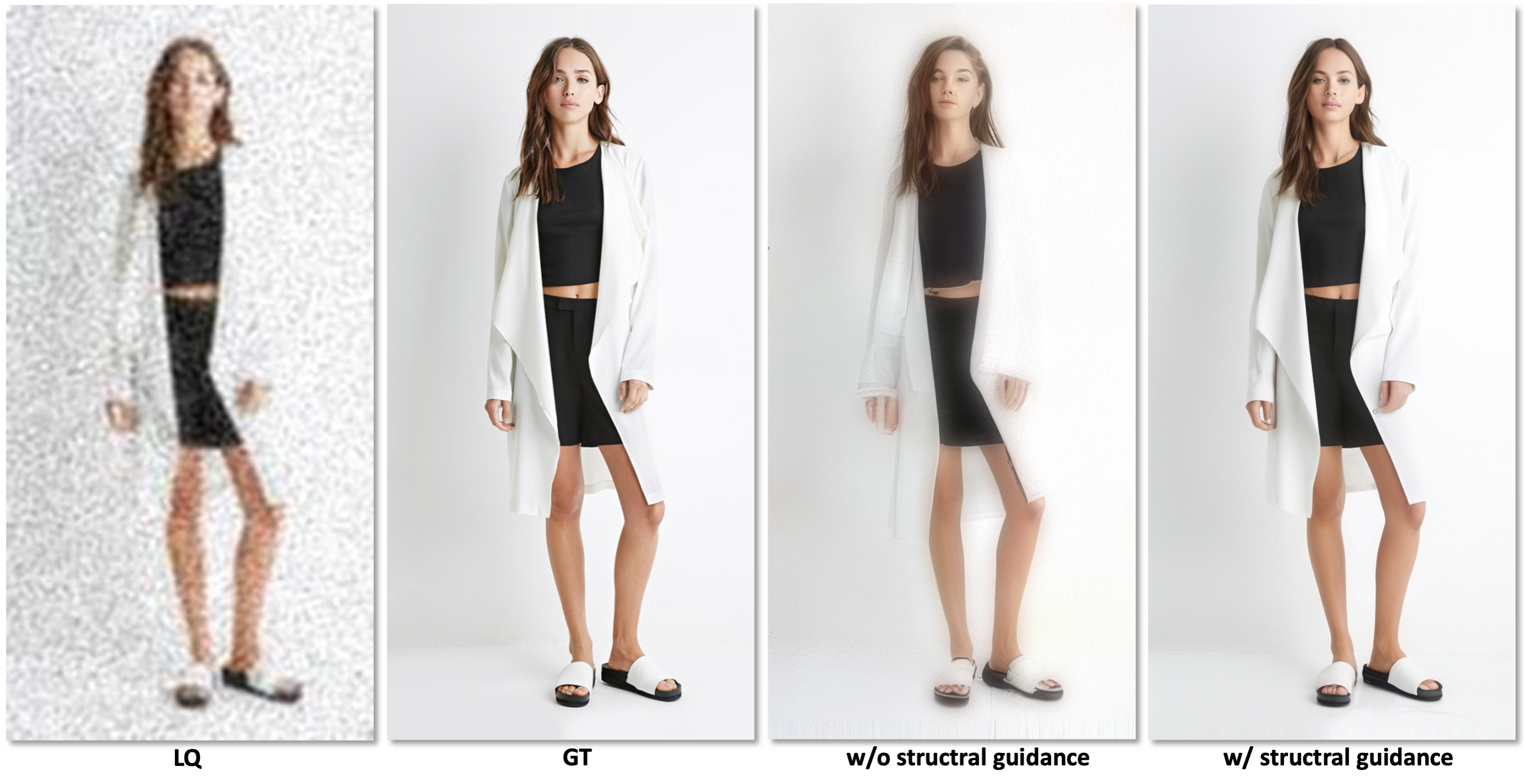}
  \label{fig:subfigp1}
\end{subfigure}%
\\
\begin{subfigure}{\linewidth}
  \centering
  \includegraphics[width=0.8\linewidth]{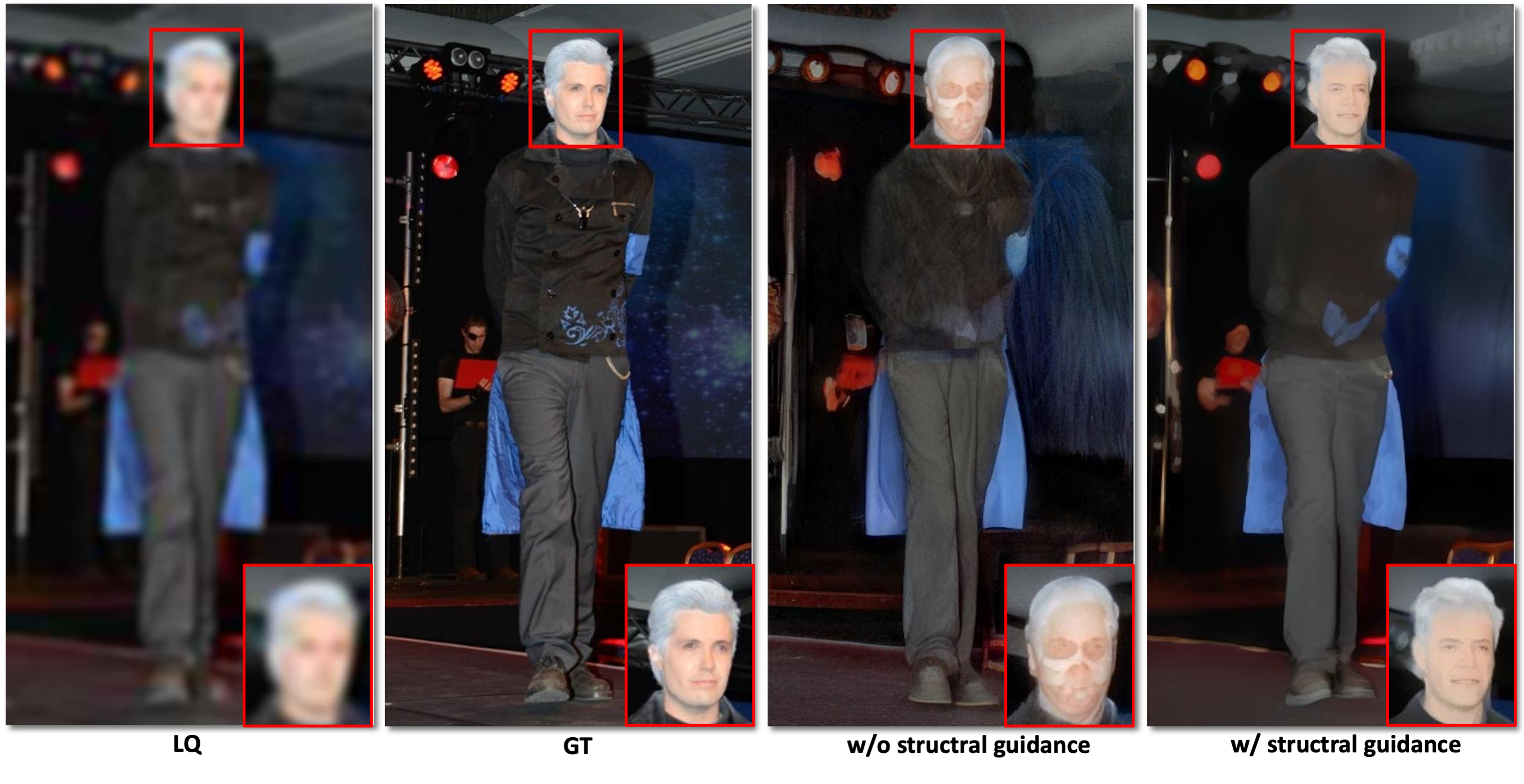}
  \label{fig:subfigp2}
\end{subfigure}
\caption{Visual comparisons of ablation study on structural guidance. With structural guidance, the problem of foreground and background blending is improved.}
\label{fig:ablation_pose_1}
\end{figure}

\vspace{-3mm}
\subsection{Ablation Studies} 
\vspace{-3mm}

\textbf{The importance of structural guidance.} In this section, we examine the importance of structural guidance within our model. By eliminating the channels dedicated to pose and attention maps, we directly fine-tune the diffusion model using identical datasets. The first row of \cref{fig:ablation_pose_1} reveals that in the absence of pose and attention map knowledge, the model struggles with foreground-background separation, especially when the color of human clothing or body parts closely resembles the background. Conversely, the second row of \cref{fig:ablation_pose_1} illustrates that incorporating pose map knowledge enables the model to more accurately "imagine" optimal facial details. These findings underscore the utility of attention and pose maps in assisting the model to accurately locate the human subject within an image, thereby enhancing the quality of human image restoration.

\begin{figure}[htbp]
\centering
\begin{subfigure}{\linewidth}
  \centering
  \includegraphics[width=\linewidth]{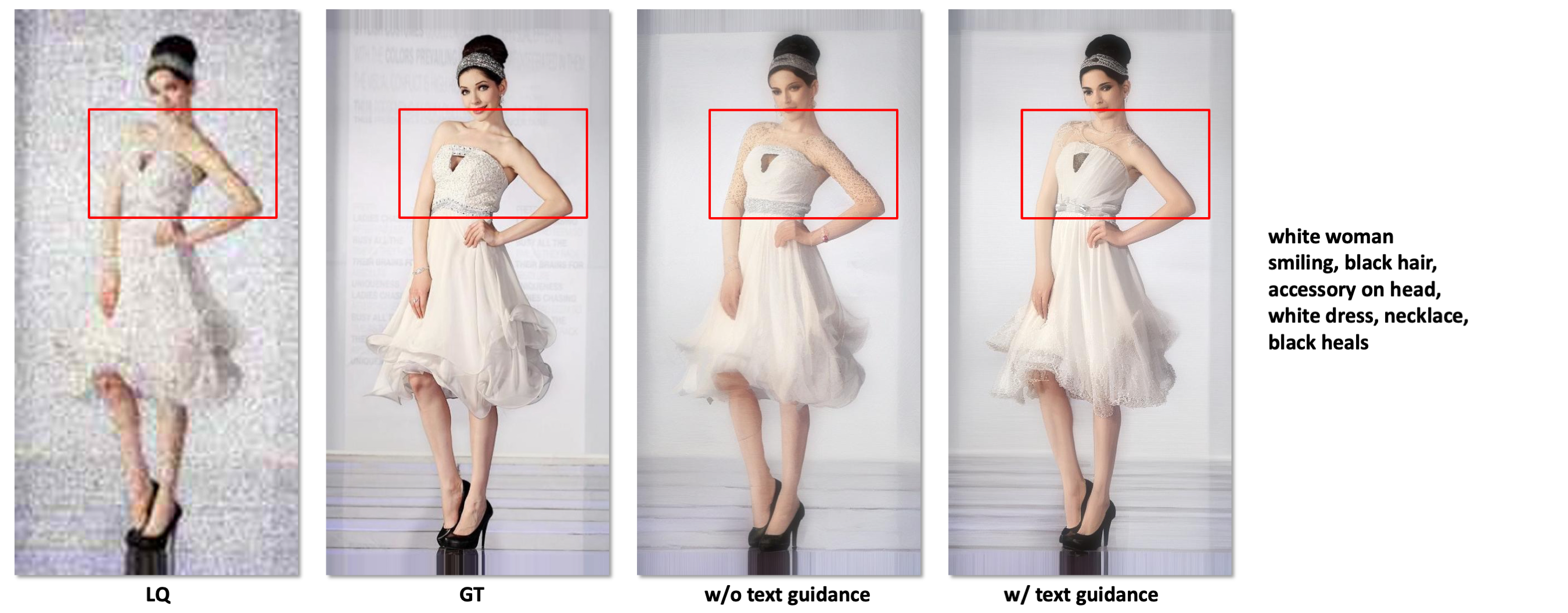}
  \label{fig:subfig1}
\end{subfigure}%
\\
\begin{subfigure}{\linewidth}
  \centering
  \includegraphics[width=\linewidth]{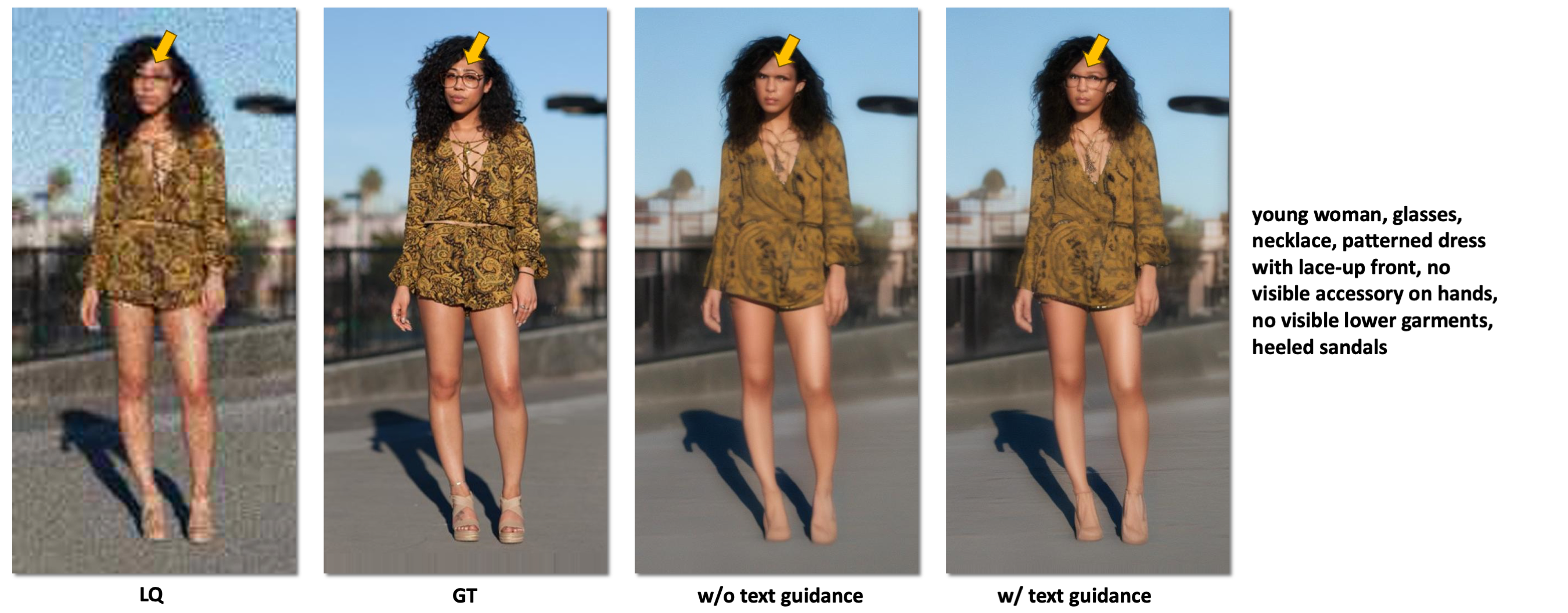}
  \label{fig:subfig2}
\end{subfigure}
\caption{Visual comparisons of ablation study on text. Texts enable the model to generate more natural human skin and improve the restoration of accessories.}
\label{fig:ablation_text_1}
\end{figure}

\vspace{-1mm}
\noindent \textbf{The importance of text guidance.} In this section, we delve into the efficacy of textual cues. We evaluate our model's performance by comparing results with and without prompts provided to our DiffBody model. The first row of \cref{fig:ablation_text_1} demonstrates that textual assistance enables DiffBody to reconstruct more lifelike human skin textures. Similarly, the second row of \cref{fig:ablation_text_1} shows that textual input allows DiffBody to accurately restore a woman's glasses—a detail challenging to recover from the degraded image alone.

\begin{figure}[ht!]
\centering
\begin{subfigure}{\linewidth}
  \centering
  \includegraphics[width=0.8\linewidth]{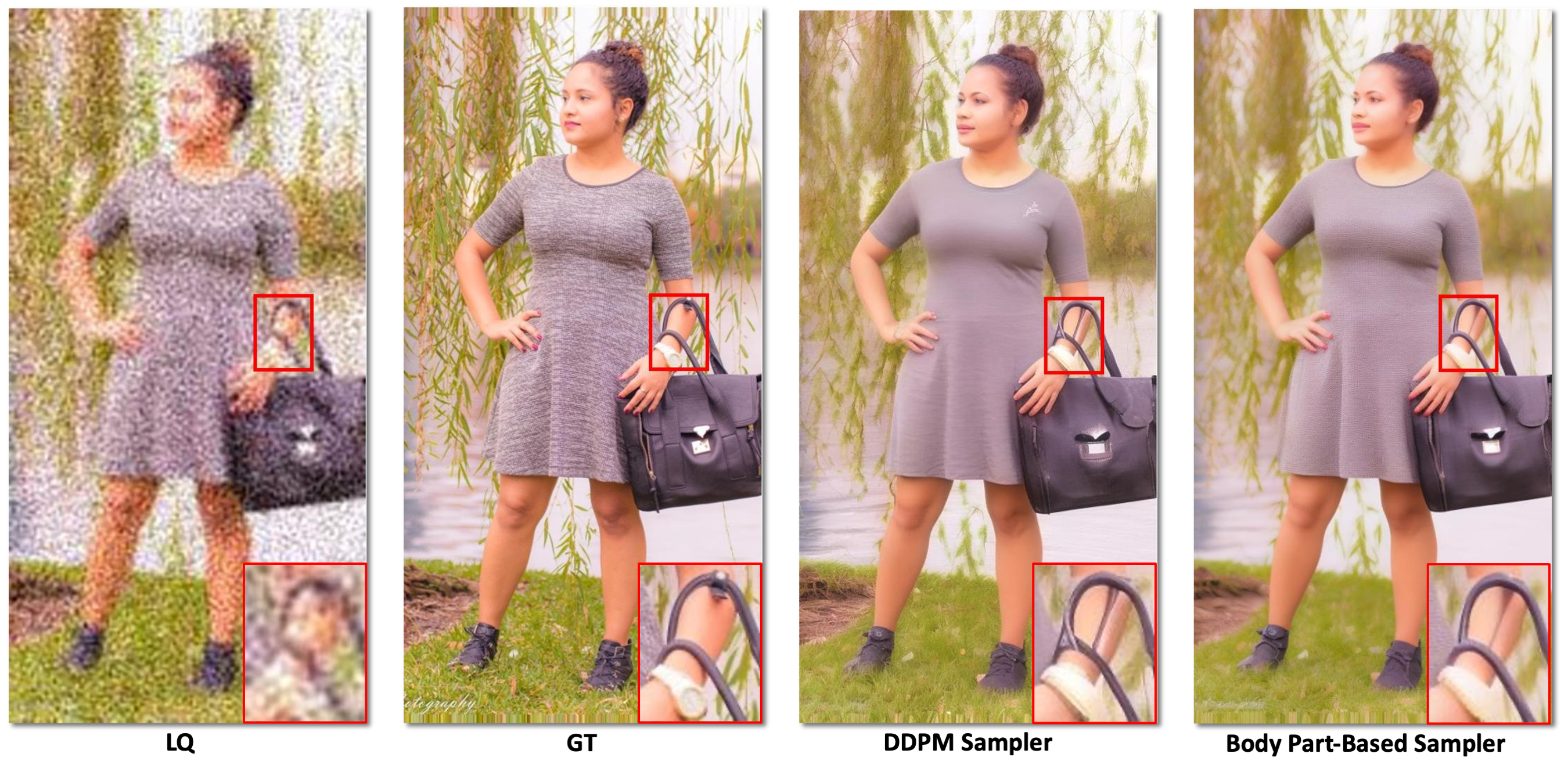}
  \label{fig:subfigs1}
\end{subfigure}%
\\
\begin{subfigure}{\linewidth}
  \centering
  \includegraphics[width=0.8\linewidth]{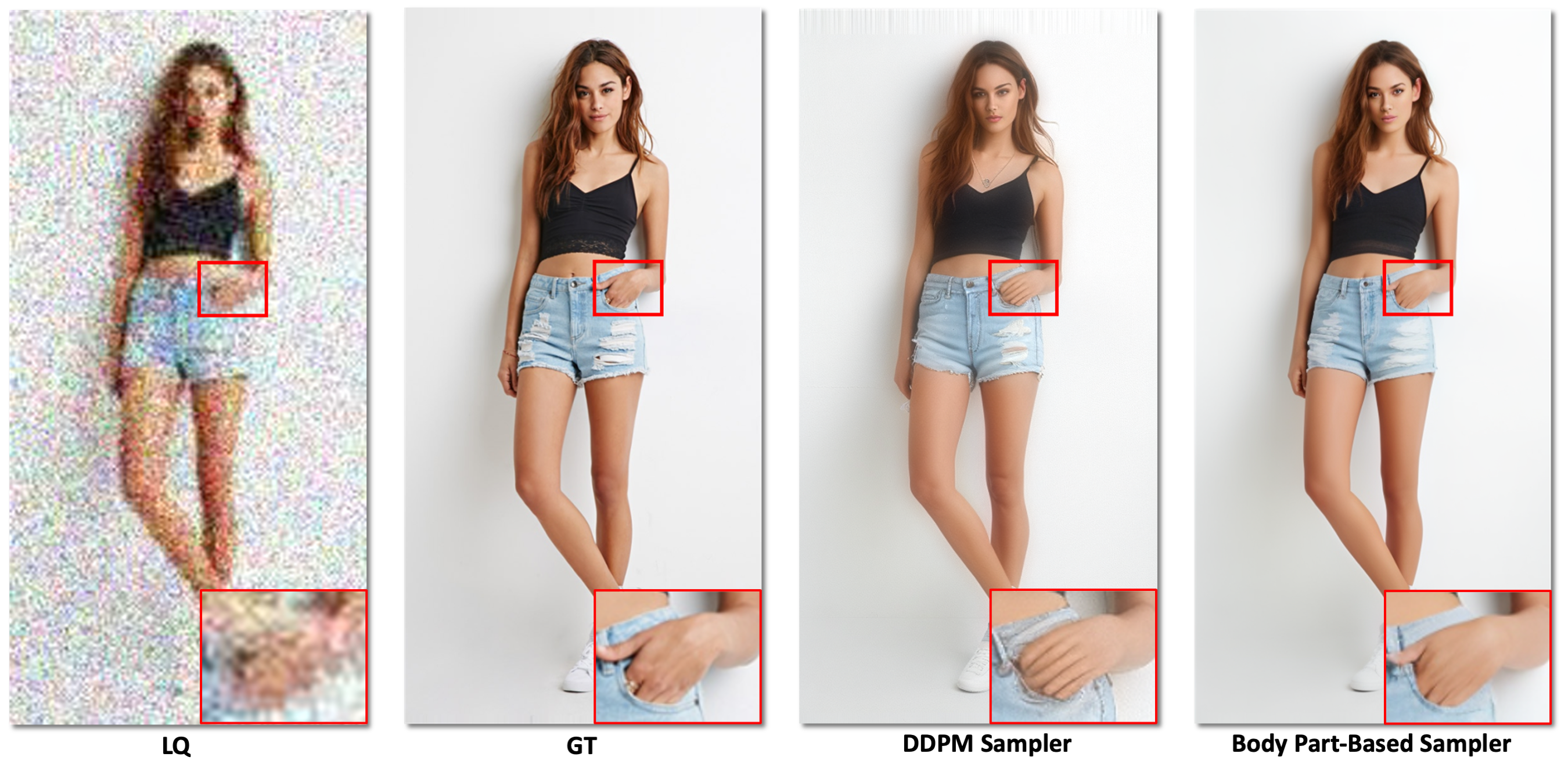}
  \label{fig:subfigs2}
\end{subfigure}
\caption{Visual comparisons of ablation study on sampler. Body part-based sampler enables the model to generate more natural limbs and fingers.}
\label{fig:ablation_sampler_1}
\end{figure}

\vspace{-1mm}
\noindent \textbf{The importance of body part-based sampler.} In this analysis, we evaluate the performance of our body part-based sampler in contrast to results obtained using the DDPM sampler. Evidence from \cref{fig:ablation_sampler_1} indicates that the body part-based sampler excels in generating more intricate limb details.

\vspace{-5mm}
\section{Conclusion}
\vspace{-3mm}

We introduce a novel framework, DiffBody, for human body restoration that achieves realistic outcomes by incorporating human-centric guidance into the pre-trained Stable Diffusion model. Through the application of various human-centric conditions, we address and rectify artifacts in human body restoration, surpassing the capabilities of existing general image restoration models. Despite DiffBody's impressive performance, the framework has yet to fully investigate advanced human body manipulation techniques, such as mesh modeling \cite{bogo2016keep}. Moreover, the preservation of personal identity within the restoration process, a crucial aspect of human image restoration, remains unexplored. We advocate for further research into leveraging Stable Diffusion for enhanced human image restoration, emphasizing the need for sophisticated body control and identity preservation.

\title{Supplementary Material}

\titlerunning{DiffBody}

\author{}

\authorrunning{Zhang et al.}
\institute{}

\maketitle
\section{Detailed Illustration on GPT-4V Prompt}

Below is a detailed prompt we provide to GPT-4V to caption our dataset. These sentences are provided to GPT-4V simultaneously as a prompt list. Such prompts ensure GPT-4V captions our dataset with texts in the same format.
\begin{figure}
    \centering
    \includegraphics[width=0.9\linewidth]{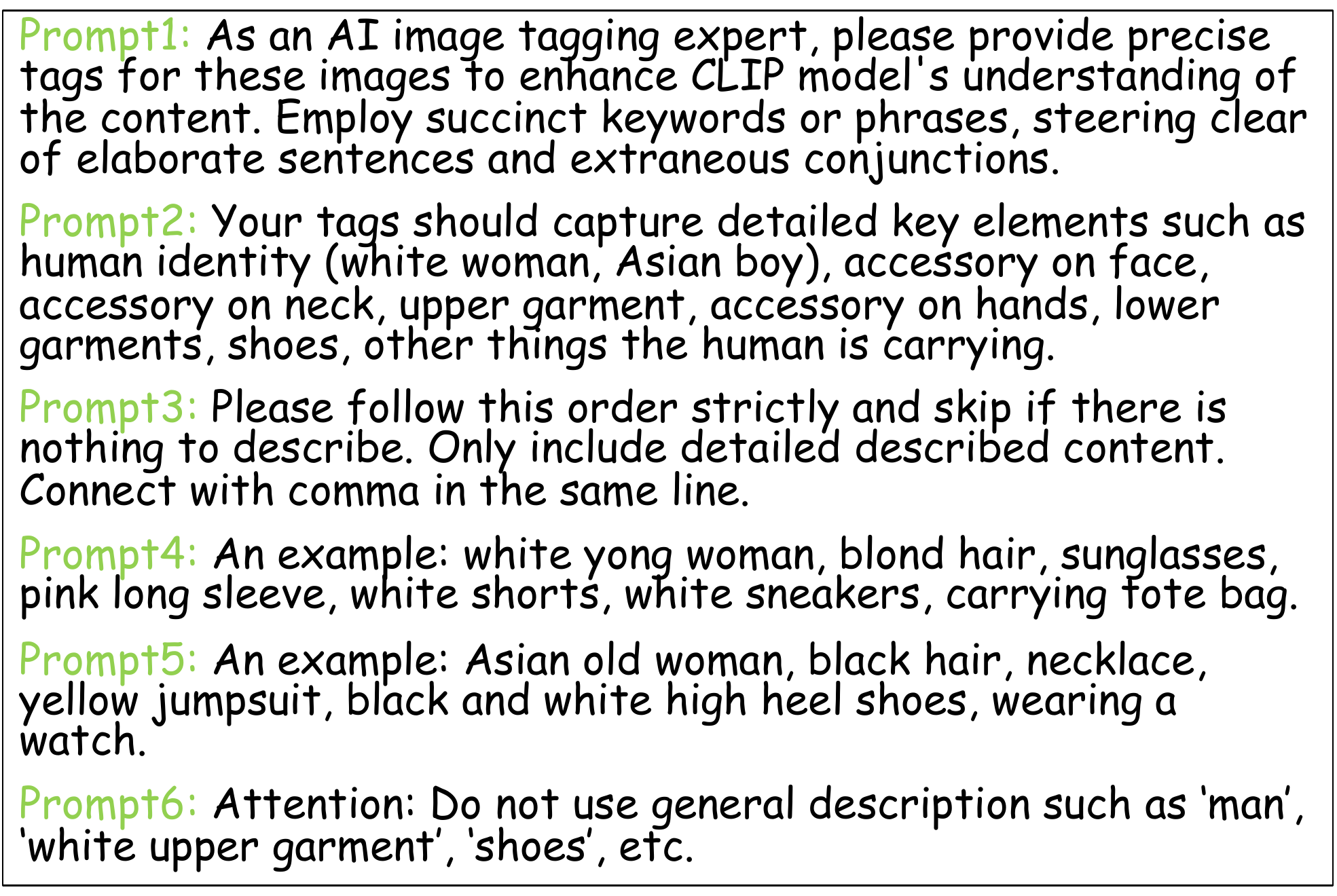}
    \caption{Detailed prompt we provide to GPT-4V to caption our dataset.}
    \label{fig:enter-label}
\end{figure}
\section{More Qualitative Comparisons}

\begin{figure}[htbp]
    \centering
    \captionsetup{skip=5pt}
    \includegraphics[width=\textwidth]{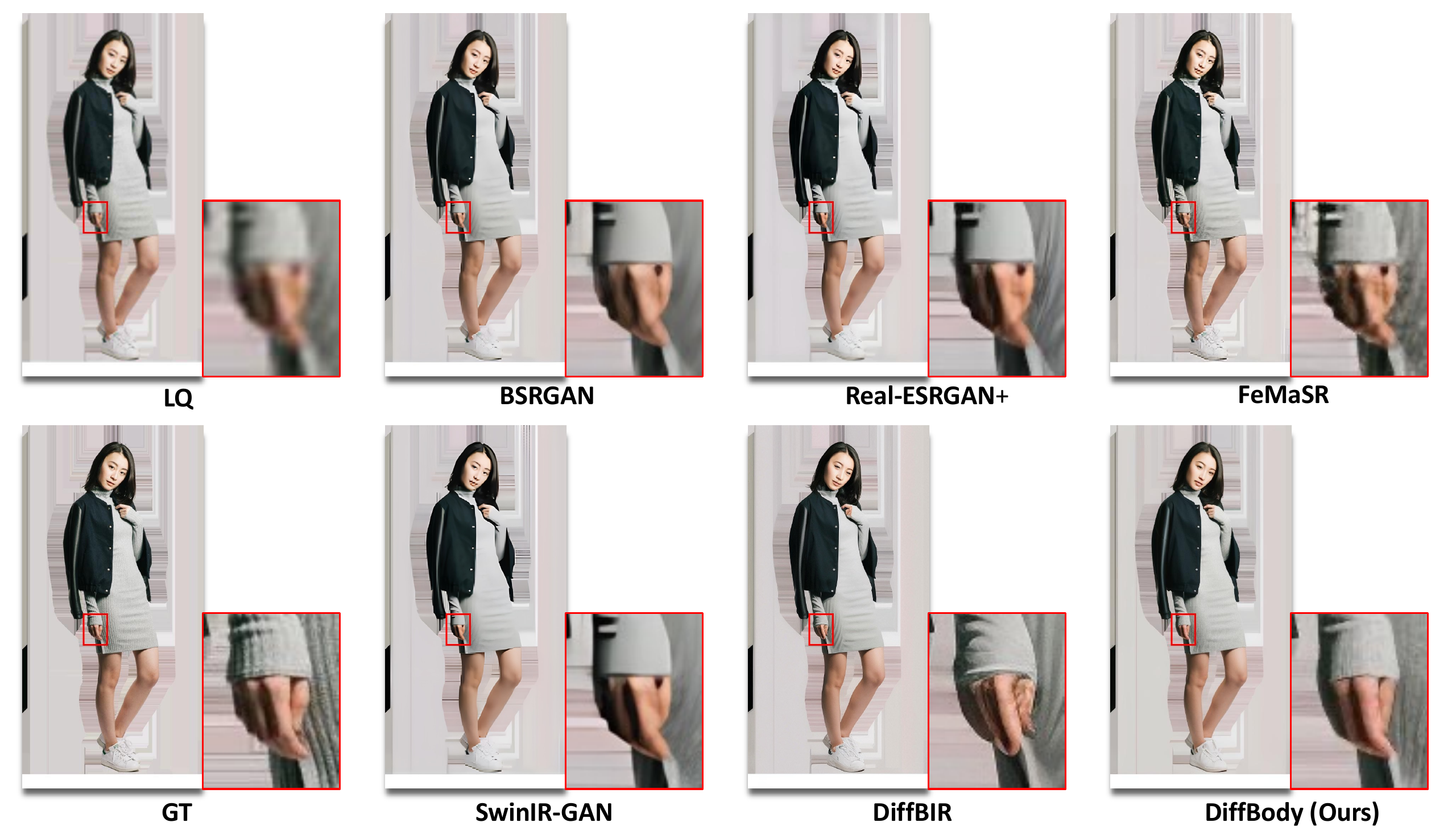}
    \caption{Visual comparison of DiffBody and other general SOTA methods. Compared to other methods, our model is more effective in generating detailed limbs.}
    \label{fig:sota11}
\end{figure}

\begin{figure}[htbp]
    \centering
    \captionsetup{skip=5pt}
    \includegraphics[width=\textwidth]{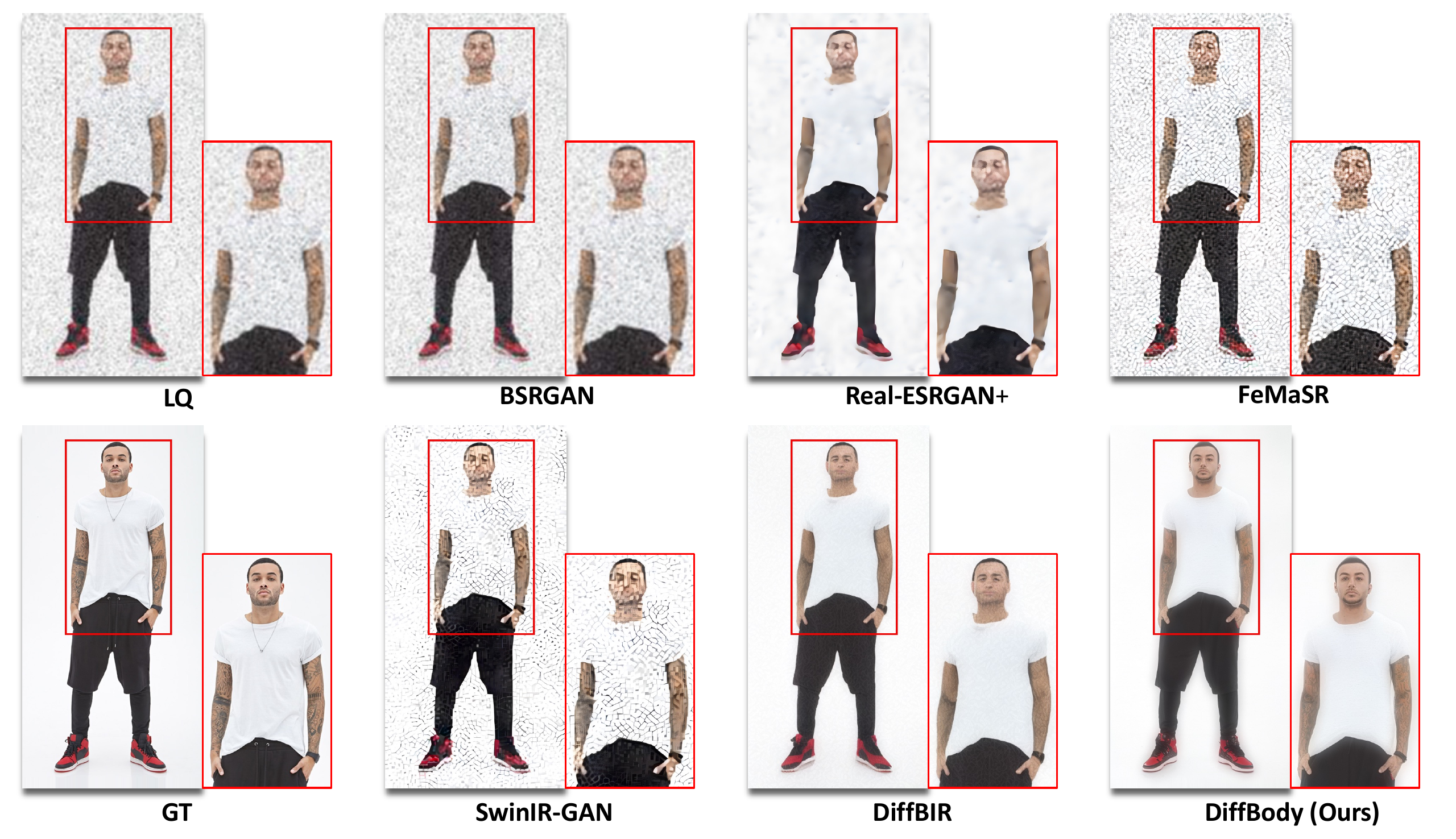}
    \caption{Visual comparison of DiffBody and other general SOTA methods. Compared to other methods, our model is more effective in generating reasonable human faces and skin texture.}
    \label{fig:sota22}
\end{figure}

\begin{figure}[htbp]
    \centering
    \captionsetup{skip=5pt}
    \includegraphics[width=\textwidth]{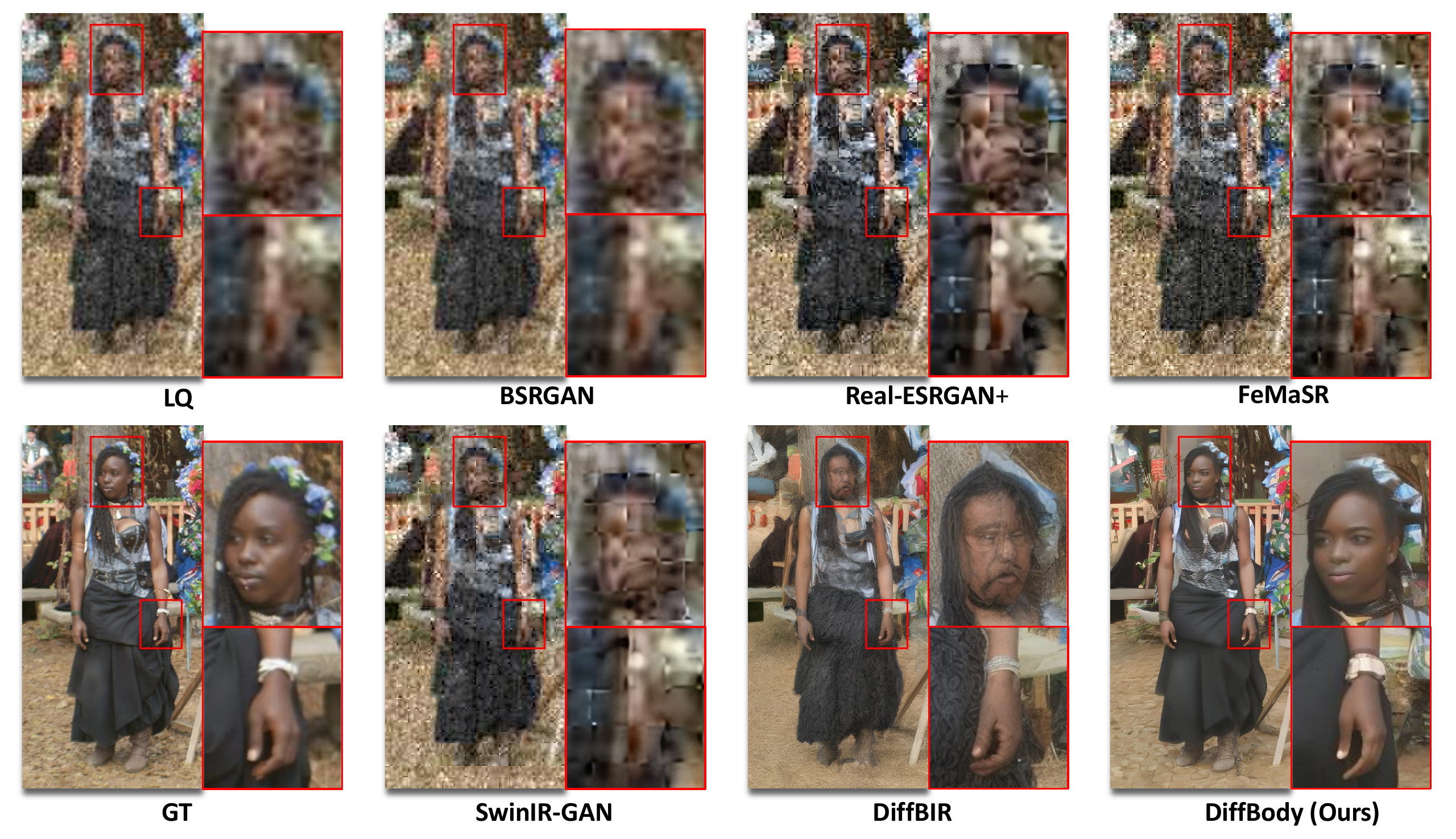}
    \caption{Visual comparison of DiffBody and other methods under extremely degraded cases (coupled degradation including adding noise, blur, and JPEG compression).}
    \label{fig:sota33}
\end{figure}

\begin{figure}[htbp]
    \centering
    \captionsetup{skip=5pt}
    \includegraphics[width=\textwidth]{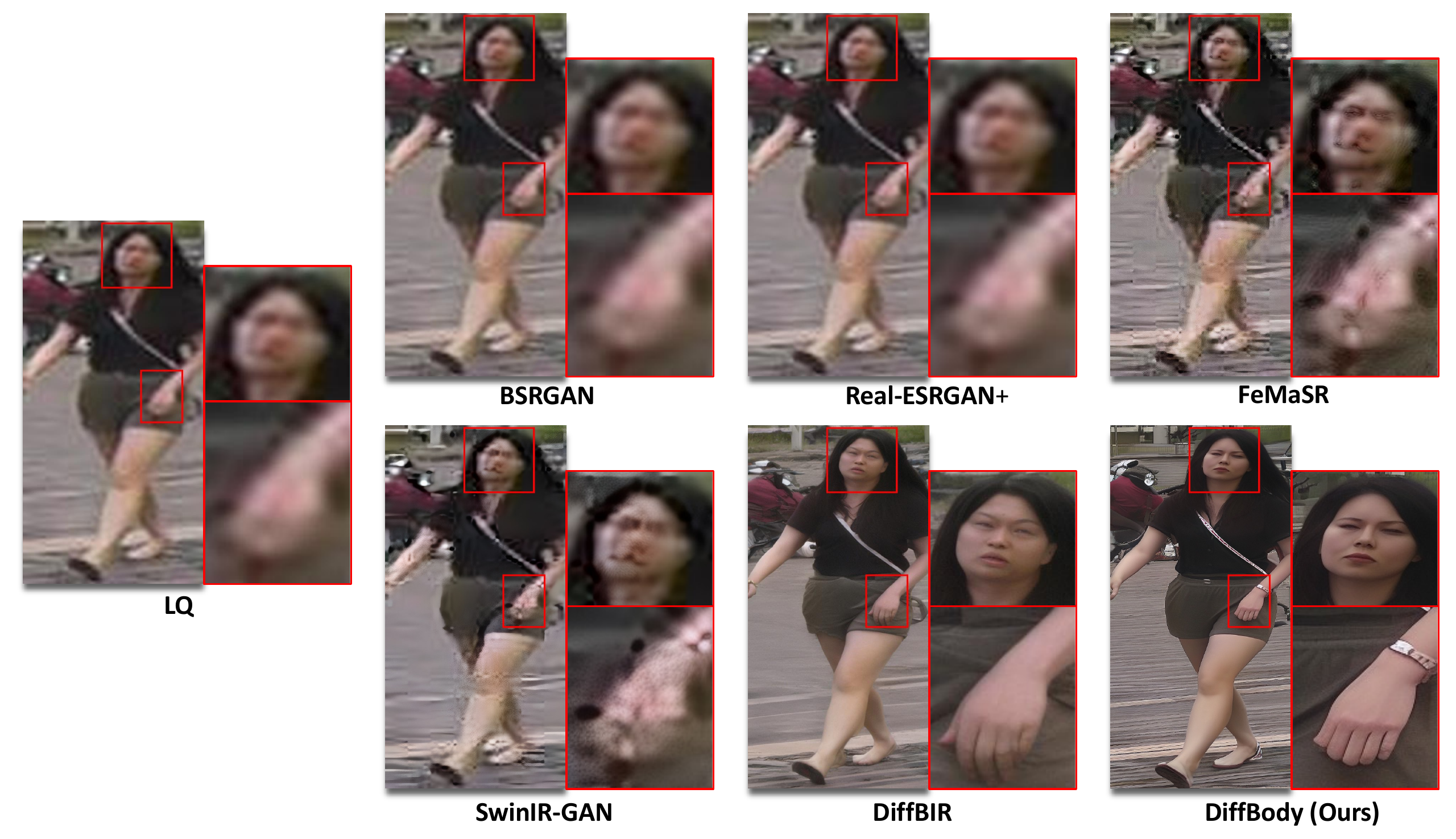}
    \caption{Visual comparison of DiffBody and other general SOTA methods on real world images. Compared to other methods, our model is more effective in generating facial and limb details.}
    \label{fig:sota44}
\end{figure}

\begin{figure}[htbp]
    \centering
    \captionsetup{skip=5pt}
    \includegraphics[width=\textwidth]{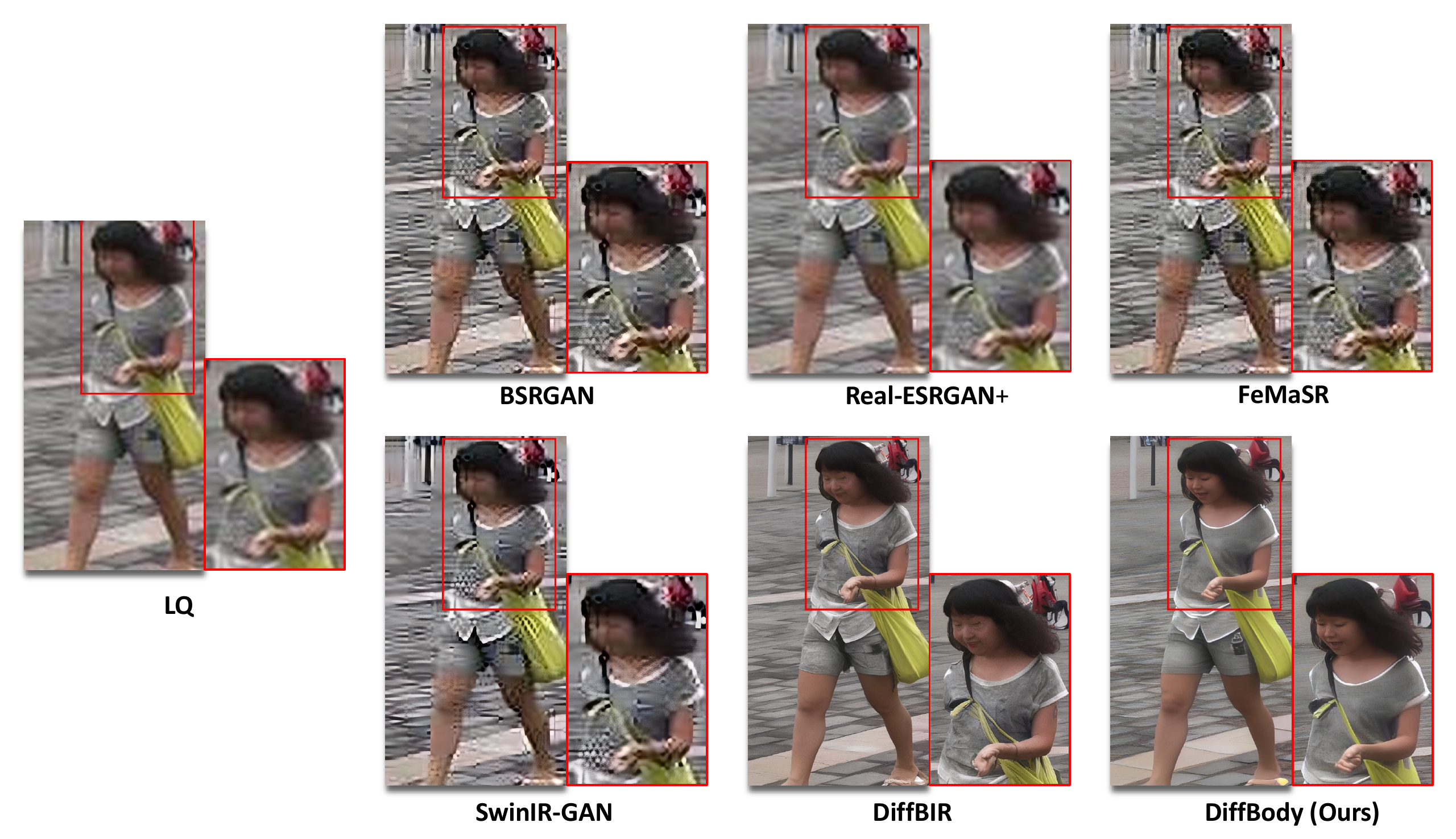}
    \caption{Visual comparison of DiffBody and other general SOTA methods on real world images. Compared to other methods, our model is more effective in generating facial and limb details.}
    \label{fig:sota55}
\end{figure}

\begin{figure}[htbp]
    \centering
    \captionsetup{skip=5pt}
    \includegraphics[width=0.77\textwidth]{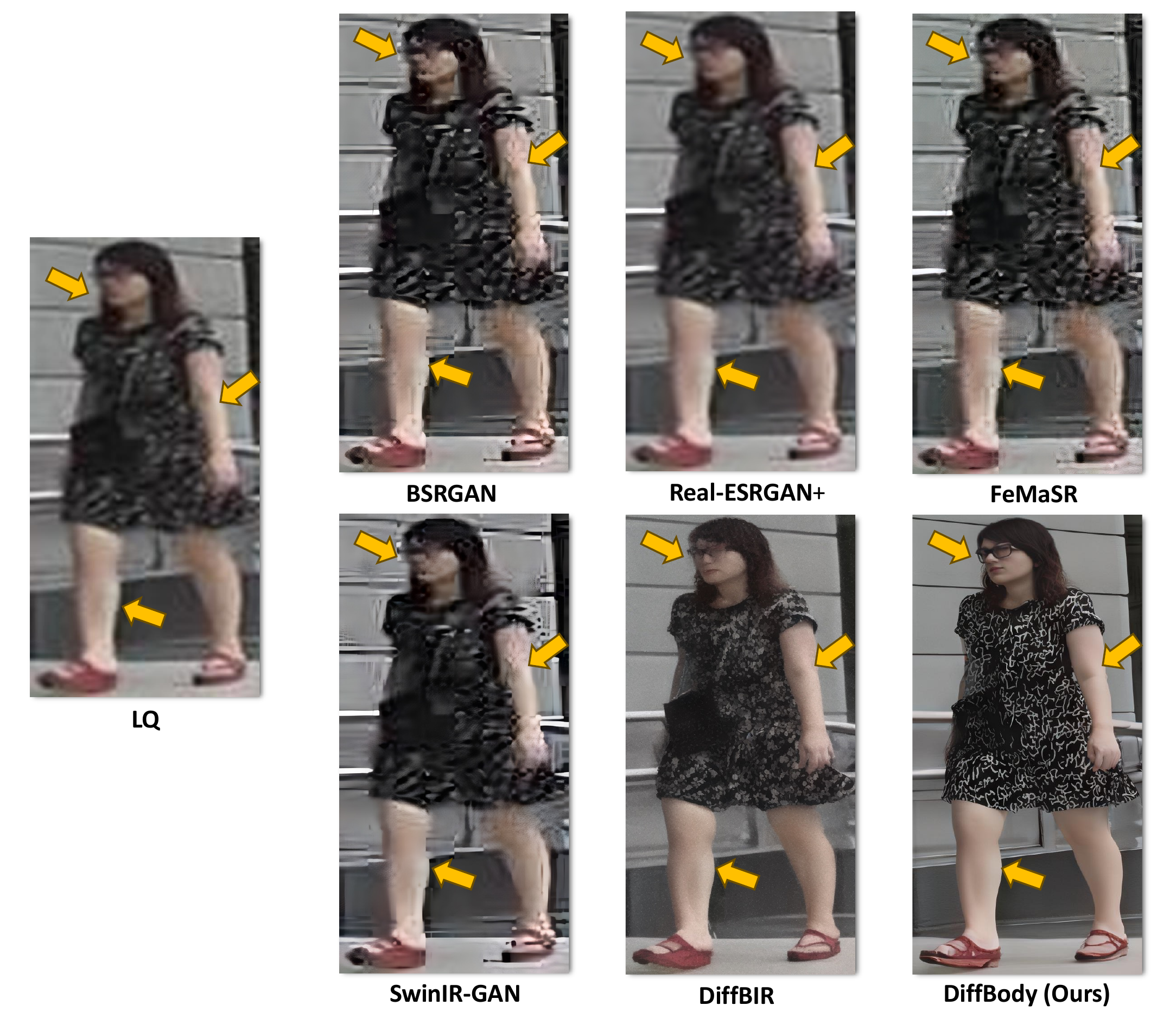}
    \caption{Visual comparison of DiffBody and other general SOTA methods on real world images. Compared to other methods, our model is able to generate more natural human skin and improve the restoration of accessories. (Zoom in for more details)}
    \label{fig:sota66}
\end{figure}

Here we add more comparisons of DiffBody with other leading general models. We also compare the results on real-world low-quality human images from the Market-1501 dataset, which contains noisy human images with a resolution of 128 × 64. In \cref{fig:sota11}, we can see that DiffBody exhibits superior facial and limb detail restoration capabilities to other methods. \cref{fig:sota22} shows that our model is more powerful in reconstructing human body textures. \cref{fig:sota33} shows that under severely degraded conditions, only DiffBody is able to reconstruct the human portrait successfully. \cref{fig:sota44} - \cref{fig:sota66} are the comparisons done on real-world low-quality human images. We can see that our model performs better than other methods on facial and limb details.



\section{User Study}

\begin{figure}[h!]
    \centering
    \captionsetup{skip=5pt}
    \includegraphics[width=0.8\textwidth]{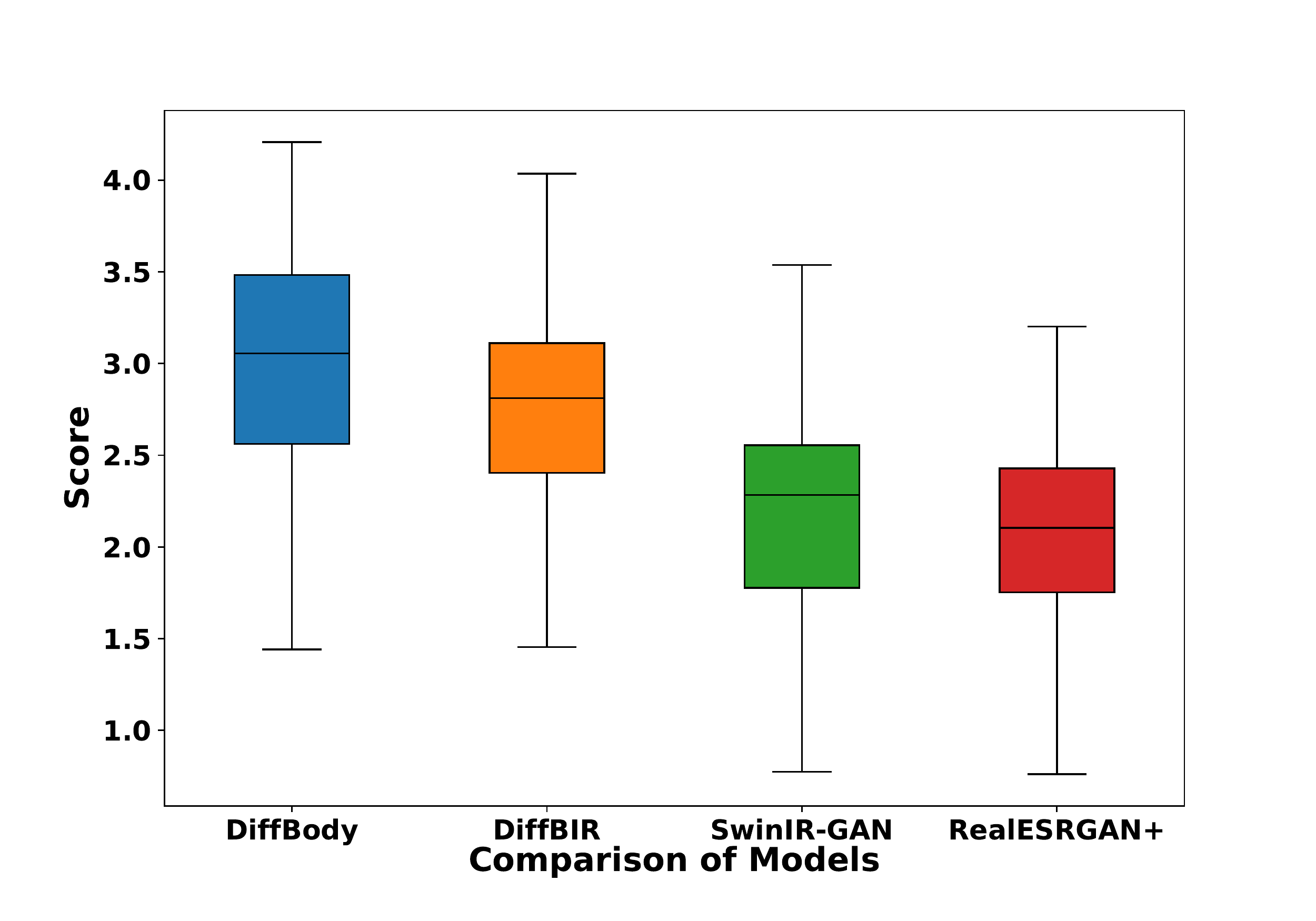}
    \caption{The distribution of scores of RealESRGAN+, SwinIR-GAN, DiffBIR, and DiffBody in user study.}
    \label{fig:user}
\end{figure}

\vspace{-3mm}
We conducted a comparative user study involving 50 images selected from our test dataset. The purpose of this study was to evaluate and compare the performance of four advanced image enhancement models: DiffBody, DiffBIR, SwinIR-GAN, and RealESRGAN+. Participants in the study were asked to review the output from each model for every image and assign a score ranging from 0 to 5, with higher scores indicating superior image quality. A group of 12 evaluators was engaged for this analysis. The collected scores for each model are illustrated in \cref{fig:user}. The results demonstrate that DiffBody outperforms the competing models, as evidenced by its highest average scores. This finding is consistent with the superior performance of DiffBody on both the MANIQA and CLIPIQA image quality assessment metrics, substantiating its efficacy in generating visually appealing results compared to the alternatives.
\newpage
\bibliographystyle{splncs04}
\bibliography{egbib}
\end{document}